\documentclass{article}
\usepackage{spconf,amsmath,epsfig}
\usepackage{xcolor, soul}
\sethlcolor{yellow}

\usepackage{multicol}
\usepackage{subfigure}
\usepackage{float}
\usepackage{hyperref}
\title{Rapid Wildfire Hotspot Detection Using Self-Supervised Learning on Temporal Remote Sensing Data}
\name
  {Luca Barco$^{1,2}$, 
  Angelica Urbanelli$^{1}$,
  Claudio Rossi$^{1}$}
  
\address{
    1. LINKS Foundation, \textit{AI, Data \& Space (ADS)}, Torino (TO), Italy \\
    2. Politecnico di Torino, \textit{Dipartimento di Automatica e Informatica  (DAUIN)}, Torino (TO), Italy \\
}
\begin{document}
%
\maketitle
\def\thefootnote{\arabic{footnote}}
\def\thefootnote{}\footnotetext{This work is based on EUMETSAT MSG High Rate SEVIRI Level 1.5 Image Data 2012-2023 and it was carried out in the context of the HORIZON-CL2-2022-HERITAGE project RescueMe (GA n.101094978), HORIZON-EUSPA-2021-SPACE project OVERWATCH (GA n.101082320) and H2020 project SAFERS (GA n.869353).}\def\thefootnote{\arabic{footnote}}

\begin{abstract}
Rapid detection and well-timed intervention are essential to mitigate the impacts of wildfires. 
Leveraging remote sensed data from satellite networks and advanced AI models to automatically detect hotspots (i.e., thermal anomalies caused by active fires) is an effective way to build wildfire monitoring systems.
In this work, we propose a novel dataset containing time series of remotely sensed data related to European fire events and a Self-Supervised Learning (SSL)-based model able to analyse multi-temporal data and identify hotspots in potentially near real time. We train and evaluate the performance of our model using our dataset and Thraws, a dataset of thermal anomalies including several fire events, obtaining an F1 score of \textbf{63.58}.
\end{abstract}

\begin{keywords}
Remote sensing, computer vision, machine learning, hotspot classification, timeseries analysis.
\end{keywords}

\section{Introduction}
\label{sec:intro}
The increasing frequency and intensity of wildfires are alarming signs and extremely dangerous environmental threats. These disruptive events, not only compromise biodiversity and ecosystems, but also endanger human safety and economic resources. Effective early detection and prompt intervention are crucial for mitigating wildfire impacts, particularly under extreme weather conditions, such as high winds and elevated temperatures, which are becoming more frequent due to climate change effects. In this context, developing systems for near real-time wildfire identification through the detection of thermal anomalies, i.e. hotspots, is vital. These anomalies, which may be indicative of fire outbreaks or other phenomena like volcanic or industrial activities, allow for quicker alerts and responses, thereby enhancing firefighting efforts.

The availability of numerous satellite networks with broad coverage and frequent revisit intervals has enabled the use of remotely sensed data in conjunction with Artificial Intelligence algorithms for automatic and near real-time hotspot detection. This technology is especially essential and advantageous in vast, sparsely populated areas where traditional on-site surveillance methods are either limited or impractical.


This work aims to contribute to the development of efficient and proactive wildfire monitoring systems leveraging remote sensed data and AI. The novel contribution of this work is the creation and the open release of (i) a new dataset enabling the training of supervised machine learning models aimed at rapid detection of hotspots (ii) a benchmark deep learning model capable of performing such task within Europe using temporal sequences derived from existing satellite networks. 
The code and the dataset are public available at \url{https://github.com/links-ads/igarss-multi-temporal-hotspot-detection}.

\section{Related Works}
\label{sec:Related}
The hotspot detection task and the monitoring of active wildfires using remote sensing data pose several challenges. First, different factors can affect data availability, i.e., satellite revisit time and geospatial resolution, cloud coverage and other atmospheric conditions. Second, finding a suitable trade-off to satisfy the need to provide a service featuring both high resolution to detect small-scale events and high temporal resolution to rapidly signal the occurrence of a wildfire. Third, design a scalable solution that enables the implementation of a service capable of covering large geographic areas while minimizing operational costs. Fourth, addressing the scarcity of ad-hoc datasets, which is of paramount importance to creating AI-based models using a supervised machine learning approach. 

In our previous work \cite{urbanelli2023multimodal}, we proposed a dataset and a machine learning model for hotspot detection and classification using remote sensing data from different sources, including MODIS, VIIRS and Sentinel-3. However, we performed the task considering only input features at a single time step. In this work, we overcome this limitation by considering time series analysis. Furthermore, we consider a different satellite network featuring shorter revisit times. We aim to improve wildfire detection accuracy by capturing temporal patterns while significantly reducing the detection time.

Recently, several works explored the Self-Supervised Learning (SSL) approach applied to remote sensing data. This is a machine learning paradigm where models learn representation from the data itself without needing labelled data. It offers advantages in feature learning, representation, and transferability to downstream tasks \cite{Reed2022ScaleMAEAS, Jain2022MultimodalCL}. Also, specific datasets have been created for this task \cite{wang2023ssl4eos12} and ad-hoc architectures have been designed, i.e., Presto \cite{tseng2023lightweight}, an SSL-model specifically designed to work with time-series of remote sensing data. It is a masked autoencoder that processes \textit{pixel timeseries}, i.e., timeseries composed of single pixels from Sentinel-1 and Sentinel-2. The training strategy is based on masking data and asking the model to reconstruct them. The model takes as input also other information that can provide additional context to the data: the land cover, the geographical coordinates of pixels timeseries and the month of the year. The Sentinel channels are processed in groups based on their central wavelength. This is useful to recognize potential similar patterns and characteristics and extract meaningful features from similar channels. 

\section{Dataset}
\label{sec:Dataset}

\begin{figure}[ht!]
    \centering
    
    \includegraphics[width=.48\textwidth]{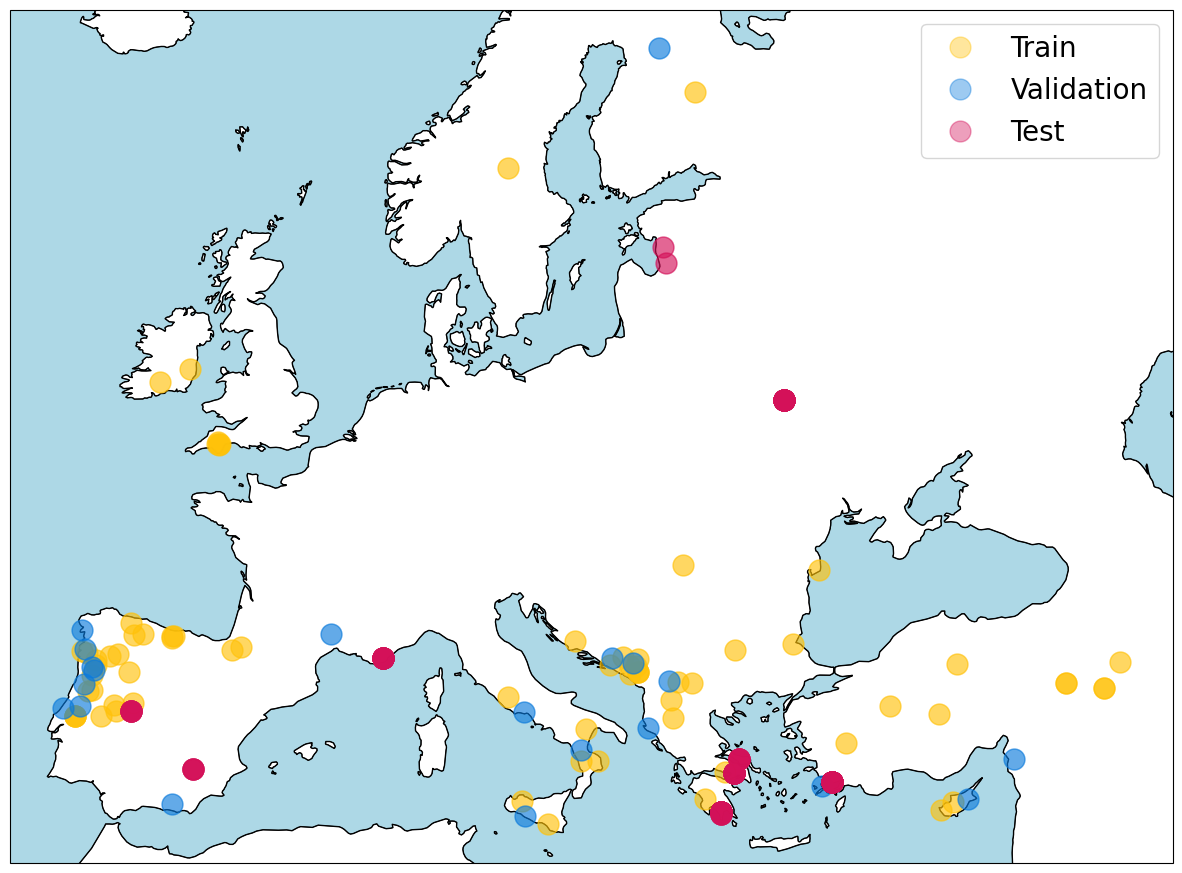}
    \caption{Geographical distribution of train, validation and test split of our dataset.}
    \label{fig:map}
    \end{figure}

In this work, we leverage three reliable data sources to address the challenges associated with the hotspot detection task and to construct a new dataset covering the Europe.
\begin{figure}[t!]
    
    \includegraphics[width=.48\textwidth]{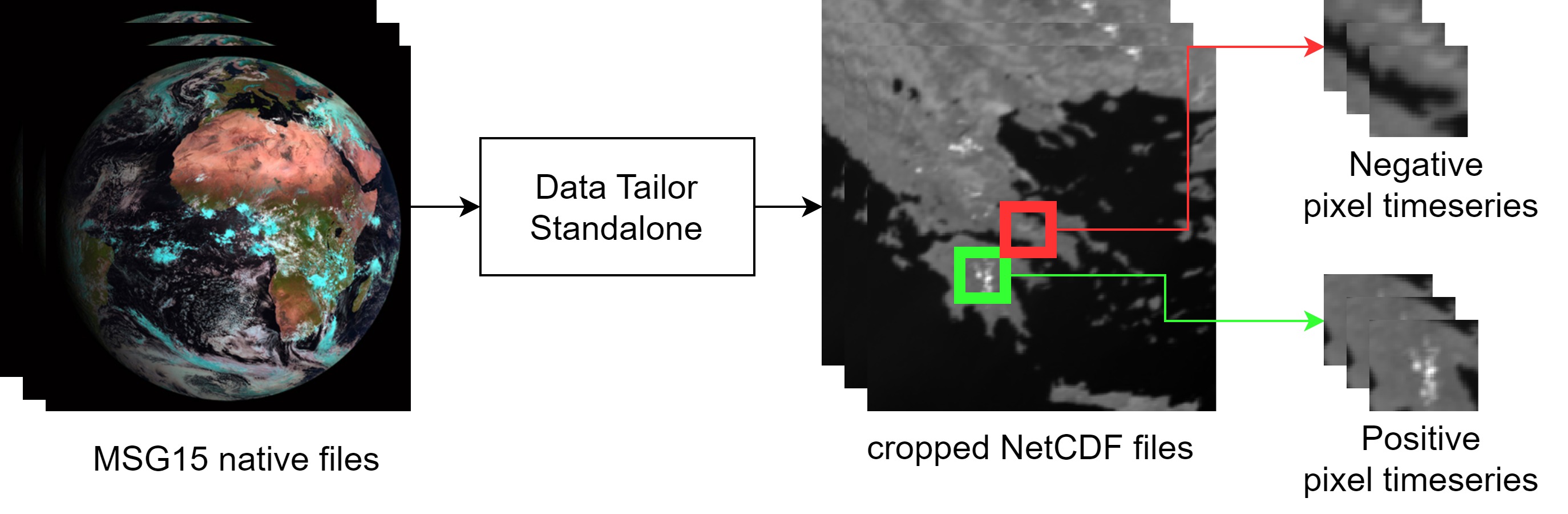}
    \label{fig:pix}
    \caption{How we build positive (greed) and negative (red) pixel timeseries, starting from MSG15 native files.}
    
\end{figure}
The first one is the European Forest Fire Information System (EFFIS) \cite{effis}, which is an information system developed by the European Commission's Joint Research Centre (JRC) aimed to monitor and report on forest fires in Europe. EFFIS offers a Rapid Damage Assessment module containing a Burnt Area product derived from the daily processing of multi-source satellite imagery.

The second one is the Meteosat Second Generation (MSG) \cite{msg}, which is a series of geostationary meteorological satellites operated by the European Organization for the Exploitation of Meteorological Satellites (EUMETSAT) with the objective of providing continuous and high-resolution imagery of the Earth's atmosphere for weather monitoring and forecasting purposes. Specifically, we use the MSG High Rate SEVIRI (HRSEVIRI) Level 1.5 Image Data, which provides data with 11 spectral bands covering infrared, visible and water vapour with different central wavelength. The sensors provide both a full scan every 15 minutes (MSG15) and a rapid scan every 5 minutes (MSG5) covering a latitude range from approximately 15° to 70° with a resolution up to 4km per pixel. 

The last one is ESRI 10m annual Land Use Land Cover dataset (LULC)\cite{lc}. It contains 9 classes: water, trees, flooded vegetation, crops, built area, bare ground, snow/ice, clouds and rangeland.

Meteosat provides the native files coming from satellites that can be processed to obtain a crop on the interested AoI in a specific file format (i.e. GeoTiff, NetCDF, etc.) using a tool called Meteosat Data Tailor \cite{dt}. This processing phase can be done using two approaches. One way is to access this service provided by Meteosat through REST APIs. The second way is to install a local instance of it, that is the mandatory approach for large amount of data. Thus, in our work, we download all the native files and then we process them using our local instance of Data Tailor to obtain NetCDF files cropped on the geographical area containing the fire. Although necessary, this process is extremely time-consuming and resource-intensive. For example, a two-day fire event requires the download of 192 native and unprocessed files from MSG15. This download lasts 7 hours, resulting in  approximately 50 GB of disk occupancy and the processing using Data Tailor requires 10 minutes. 

\begin{figure*}[ht!]
    \centering
    \includegraphics[width=.9\textwidth]{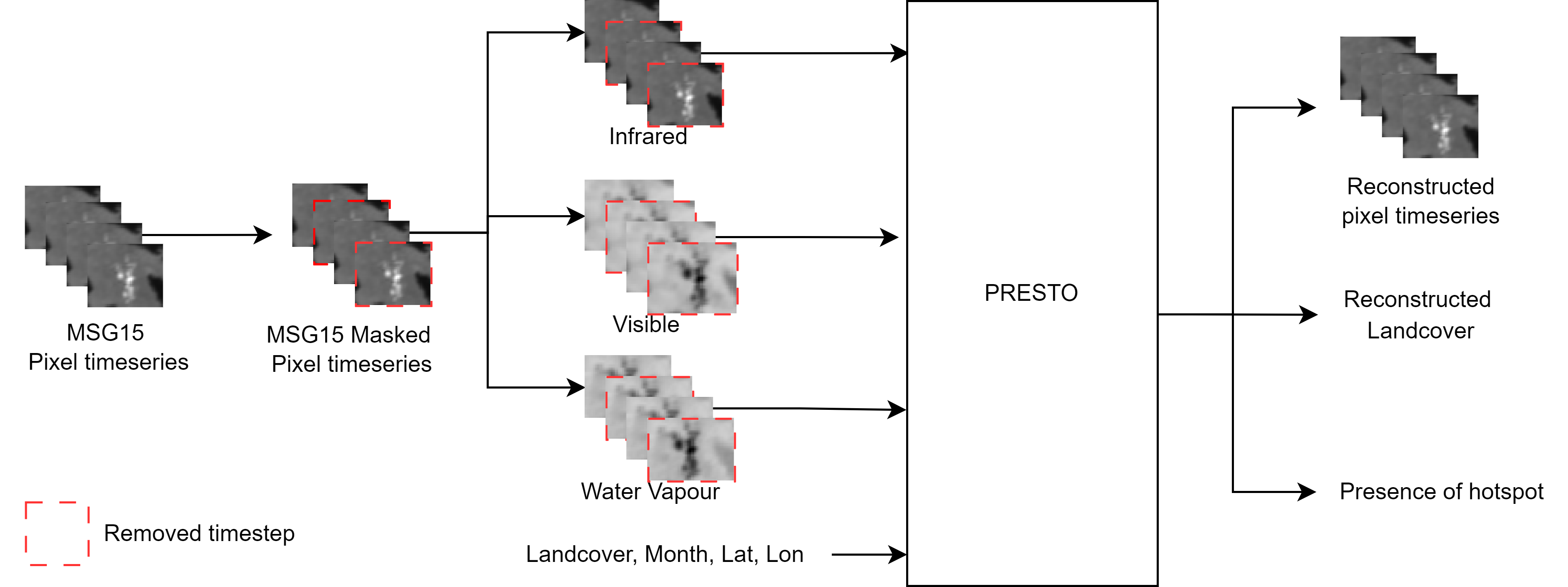}
    \caption{Architecture of the proposed approach.}
\label{fig:arch}
\end{figure*}

Our dataset is derived from fire events detected by EFFIS between 2012 and 2022. From this list, we randomly select a subset of 166 events. The decision to limit the dataset rather than using the complete list of events is motivated by all the considerations presented above about downlaod time and files size. Each fire event is identified by an Area of Interest (AoI), a start date and an end date. We download and produce the NetCDF files using the approach explained above. Then, we extract the pixels that intersects the AoI of the event and we stack them together along the temporal dimension, obtaining pixel timeseries labelled as positive. We also extract other pixels outside the AoI and label them as negative to produce pixel timeseries that don't contain hotpots. Figure \ref{fig:pix} contains an overall schema of all these steps.
Furthermore, to make the model more robust, for each pixel we also add the relative land cover class from ESRI LULC.

Finally, we obtain a dataset containing a list of positive and negative pixel timeseries. Then, we split it into train, validation and test components, proportionally satisfying the split by event over the latitude and longitude ranges to improve the model generalization capability. 
The test split has been guided by Thraws\cite{meoni2023thraws}, a dataset that includes thermal anomalies related to 88 events (20 fire events, 58 volcanic eruptions and 10 not-events) and Sentinel-2 data, which we replace with MSG data. Since this dataset doesn't contain information about the fire start date and end date, we select from EFFIS the events contained in Thraws, so that we can use the EFFIS time range to label the pixel timeseries. Unfortunately, only 11 Thraws fire events are present in EFFIS. 
Finally, our splits contain 122 events in train, 21 events in validation and 11 events in test.
Figure \ref{fig:map} shows their geographical distribution.

\section{Methodology}
\label{sec:Material}

The proposed solution for the hotspot detection task on multitemporal remote sensing data leverages the SSL paradigm, which is subsequently applied to a binary classification task. We introduce a modified version of Presto \cite{tseng2023lightweight}, introduced in Section \ref{sec:Related}. 
In our adaptation, we replace the Sentinel-1 and Sentinel-2 inputs with MSG15 data, while retaining all the other features as specified in the original Presto paper since they are used to compute sinusoidal embeddings. 
Then, MSG15 channels are aggregated with near-central wavelengths according to Presto approach, defining three groups of channels: infrared wavelengths (channels 1, 2, 3, 4, 5, 6, 7) , visible wavelengths (channels 8, 9), and water vapor (channels 10, 11).
In summary, the model takes as input a pixel timeseries of shape $(B, TS, C)$, where $B$ is the batch size, $TS$ is the number of timesteps and $C$ is the number of channels.
Figure \ref{fig:arch} illustrates the architecture of our presented model. For training, we employ a composite loss function comprising three components: $L_{eo}$ (Mean Squared Error) for the reconstruction of time series, $L_{lc}$ (Cross Entropy Loss) for land cover reconstruction, and $L_{cls}$ (Binary Cross Entropy Loss) for hotspot classification. The total loss function is obtained by summing the three losses: 
\begin{equation}
  L_{tot} = L_{eo} + L_{lc} + L_{cls}.
\end{equation}

\section{Experiments and results}
\label{sec:Results}
In our implementation, the input size of pixel timeseries is $(BS, 96, 11)$. We set $C$ equal to the number of MSG15 channels. We set $TS = 96$ since it is the length of a 24-hours timeseries. If a timeseries is longer than 24 hours, it is split in sub-timeseries with length equal to 96. A consistent mask ratio of 0.75 is applied, with random timesteps masked in each experiment.

All experiments are conducted on a NVIDIA GeForce RTX 2080 for 100 epochs, with a learning rate of 1e-03, a batch size of 32, the Adam optimizer, and several schedulers, i.e., Step LR, Linear, Cosine, Cosine Warmup.  
For each scheduler, we run three experiments using different seeds (17, 42, 91). The scores of these three experiments are averaged and reported along with standard deviation. 

Table \ref{tab:results} contains the results in terms of F1 score of our experiments using different schedulers for both validation and test splits of our dataset in order to have a score over a random selection of the dataset (validation split) and over the events contained in Thraws (test split).

\begin{table}[h]
\begin{tabular}{|l|l|l|}
\hline
\textbf{Scheduler}     & \textbf{F1 Validation} & \textbf{F1 Test}       \\ \hline
\textit{Step LR}       & $48.54 \pm 3.41$          & $52.15 \pm 2.78$          \\ \hline
\textit{Linear}        & $\mathbf{60.57 \pm 1.71}$ & $60.49 \pm 1.71$          \\ \hline
\textit{Cosine}        & $60.27 \pm 2.22 $         & $\mathbf{63.58 \pm 0.71}$ \\ \hline
\textit{Cosine Warmup} & $50.85 \pm 0.88$          & $53.78 \pm 2.05$          \\ \hline
\end{tabular}
\caption{Experiments results using different schedulers. The Validation and Test F1 scores are reported along with the standard deviation, based on three experiments conducted with different seeds (17, 42, 91).}
\label{tab:results}
\end{table}

The results indicate that the \textit{Linear} scheduler achieved the highest F1 score on the validation set, with a mean of $60.57 \pm 1.71$, while the \textit{Cosine} scheduler, even though slightly lower on the validation set with $60.27 \pm 2.22$, achieved the highest F1 score on the test set, with a mean of $63.58 \pm 0.71$. This suggests that the \textit{Linear} scheduler is the most effective during the validation phase, however the \textit{Cosine} scheduler provides the best generalization performance on the test data.

The \textit{Step LR} scheduler results in the lowest F1 scores on both the validation and test sets, $48.54 \pm 3.41$ and $52.15 \pm 2.78$, respectively, suggesting it is less effective in optimizing the model for this specific task.

The \textit{Cosine Warmup} scheduler, instead, shows moderate performance, with F1 scores of $50.85 \pm 0.88$ on the validation set and $53.78 \pm 2.05$ on the test set, which are better than \textit{Step LR} but not as competitive as \textit{Linear} and \textit{Cosine} schedulers.

In summary, the \textit{Linear} and \textit{Cosine} schedulers demonstrate superior performance, with the \textit{Linear} scheduler excelling in validation and the \textit{Cosine} scheduler excelling in test performance.

\section{Conclusions and future works}
\label{sec:Conclusions}
In this work, we address the task of hotspot detection using time series of remotely sensed data by adopting a self-supervised learning model to autonomously learn relevant features from the unlabeled dataset. The acquired feature extractor is then employed in a downstream hotspot detection task.
Our results demonstrate the efficacy of the self-supervised approach in capturing temporal patterns and representations of remotely sensed data. 
Indeed, by using this approach, we obtain good results in terms of F1 score. 

Moreover, the self-supervised feature extractor can be used for other downstream tasks, offering wider capabilities than machine learning models specifically trained on a given task. This adaptability makes our methodology applicable to various remote sensing scenarios and encourages the exploration of different classification tasks beyond hotspot detection.

Another valuable contribution of this work is the creation of a dedicated hotspot dataset, including three reliable data sources and enabling time series analysis. The dataset can also become a benchmark for future research in the field. 

While our findings are promising, we acknowledge ongoing challenges in the domain, including data quantity and quality, interpretability of learned features, and the need for robust evaluation metrics. Future work should address these challenges and further refine the proposed methodology by enlarging the dataset in order to cover more geographical areas and exploit the entire EFFIS fire event list.
\bibliographystyle{IEEEbib}
\bibliography{strings,refs}

\begin{thebibliography}{10}

\bibitem{urbanelli2023multimodal}
Angelica Urbanelli, Luca Barco, Edoardo Arnaudo, and Claudio Rossi,
\newblock ``A multimodal supervised machine learning approach for satellite-based wildfire identification in europe,'' 2023.

\bibitem{Reed2022ScaleMAEAS}
Colorado Reed, Ritwik Gupta, Shufan Li, Sara Brockman, Christopher Funk, Brian Clipp, Salvatore Candido, Matthew Uyttendaele, and Trevor Darrell,
\newblock ``Scale-mae: A scale-aware masked autoencoder for multiscale geospatial representation learning,''
\newblock {\em ArXiv}, vol. abs/2212.14532, 2022.

\bibitem{Jain2022MultimodalCL}
Umang Jain, Alex Wilson, and Varun Gulshan,
\newblock ``Multimodal contrastive learning for remote sensing tasks,''
\newblock {\em ArXiv}, vol. abs/2209.02329, 2022.

\bibitem{wang2023ssl4eos12}
Yi~Wang, Nassim Ait~Ali Braham, Zhitong Xiong, Chenying Liu, Conrad~M Albrecht, and Xiao~Xiang Zhu,
\newblock ``Ssl4eo-s12: A large-scale multi-modal, multi-temporal dataset for self-supervised learning in earth observation,'' 2023.

\bibitem{tseng2023lightweight}
Gabriel Tseng, Ruben Cartuyvels, Ivan Zvonkov, Mirali Purohit, David Rolnick, and Hannah Kerner,
\newblock ``Lightweight, pre-trained transformers for remote sensing timeseries,'' 2023.

\bibitem{effis}
European Union,
\newblock ``{EFFIS},'' Available online: \url{https://effis.jrc.ec.europa.eu/}.

\bibitem{msg}
European Union,
\newblock ``{Meteosat MSG},'' Available online: \url{https://effis.jrc.ec.europa.eu/}.

\bibitem{lc}
Krishna Karra, Caitlin Kontgis, Zoe Statman-Weil, Joseph~C. Mazzariello, Mark Mathis, and Steven~P. Brumby,
\newblock ``Global land use / land cover with sentinel 2 and deep learning,''
\newblock in {\em 2021 IEEE International Geoscience and Remote Sensing Symposium IGARSS}, 2021, pp. 4704--4707.

\bibitem{dt}
European Union,
\newblock ``{Meteosat Data Tailor},'' Available online: \url{https://user.eumetsat.int/resources/user-guides/data-store-detailed-guide}.

\bibitem{meoni2023thraws}
Gabriele Meoni, Roberto~Del Prete, Federico Serva, Alix~De Beussche, Olivier Colin, and Nicolas Longépé,
\newblock ``Thraws: A novel dataset for thermal hotspots detection in raw sentinel-2 data,'' 2023.

\end{thebibliography}

\end{document}